# Monitoring and Prediction of Mood in Elderly People during Daily Life Activities*

Daniel Bautista-Salinas, Joaquín Roca González, *Member, IEEE,* Inmaculada Méndez, Oscar Martinez Mozos, *Member, IEEE*

*Abstract*—We present an intelligent wearable system to monitor and predict mood states of elderly people during their daily life activities. Our system is composed of a wristband to record different physiological activities together with a mobile app for ecological momentary assessment (EMA). Machine learning is used to train a classifier to automatically predict different mood states based on the smart band only. Our approach shows promising results on mood accuracy and provides results comparable with the state of the art in the specific detection of happiness and activeness.

## I. INTRODUCTION

We are currently experiencing unprecedented levels of ageing in the population around the world. In the EU there will be around 151 million people aged over 65 years by 2060. Older citizens in the EU wish to stay in their homes for as long as possible to enjoy an active and healthy ageing as the increase of 13.14% in the number of older people living alone in the EU shows [1]. However, elderly living alone are more likely to experienced poor mental health including depression, anxiety or low self-esteem [2]. Moreover, mental health in elderly has been assigned relatively low priority, it has been poorly covered by existing health monitoring systems, and it is often considered a forgotten matter [3], [4]. These facts and the increasing difficulty to cover all demands for home assistance [5], result in an urgent need for new and innovative forms of mental health support to maintain their mental well-being.

In this paper, we focus on the automatic monitoring and prediction of mood states in elderly people during their daily life activities. Therapists and psychologist identify negative mood states as main symptoms for depression and anxiety.

Advances in intelligent mood recognition using machine learning algorithms have received great interest in the last years. Promising results have been obtained in studies in laboratory settings [6], [7]. However, the sensors used are cumbersome to be worn in real life situations. In addition, these results cannot be directly extrapolated to daily life conditions due to the high number of variables that cannot be reproduce fairly in a laboratory setting. The improvements on wearable devices have allowed taking these studies to real environments and situations. For example, the works in [8], [9] built prediction models in laboratory settings that are refined in less controlled scenarios such us health ambulatories. Although in the ambulatory testing scenario the users use comfortable devices and applications, to create the laboratory model they use sensors and techniques that are not suitable for daily life use or testing. Recently, more daily life experiments have been carried out without building a laboratory model [10], [11].

The main disadvantage of these studies is that psychological questionnaires to assess mental state are too long to be asked repeatedly during the day. The validity of these questionnaires might be lost if they are replicated in such short periods. To solve this problem, our approach uses ecological momentary assessments (EMAs) that allow us to assess the mood state with two simple questions. EMA answers are extrapolated to mood states using the mental state model proposed in [11].

## II. METHODOLOGY

Our system is composed of a wristband device that records physiological and activity data from users. In addition, we use a mobile app for EMA input. In this first prototype, data are stored and processed offline. We use the Empatica E4 (Empatica Inc., USA) wristband [12]. This wearable device can measure the blood volume pulse (BVP), the electrodermal activity (EDA) and the peripheral skin temperature. It also contains a 3-axis accelerometer, and a built-in application which calculates the heart rate (HR) and the interbeat interval (IBI) from the BVP. This wristband looks like a watch and can be worn around-the-clock.

To record the participants' mood state for training our classifier, we use the approach by R. Likamwa *et al.* [13] which is based on the mood model proposed J. A. Russell [14] (Fig. 1A). This model proposes two measurements, happiness and activeness, in a 5-point Likert scale, ranging from 0 to 4. The final mood state is obtained by interpolating the happiness and activeness levels as shown in (Fig. 1A).

EMA is used to prompt users about these measurements using a mobile app that we have developed (Fig. 1B). Users are asked only 5 times per day to avoid creating extra anxiety due to a high load of questions [13]. The interface is kept

* **This is the authors' manuscript. The final published article is available at https://doi.org/10.1109/EMBC.2019.8857847**

© 2019 IEEE. Personal use of this material is permitted. Permission from IEEE must be obtained for all other uses, in any current or future media, including reprinting/republishing this material for advertising or promotional purposes, creating new collective works, for resale or redistribution to servers or lists, or reuse of any copyrighted component of this work in other works.

Research supported by Spanish Ramon y Cajal program (ref. RYC-2014-15029) and by Spanish Fundacion Seneca (ref. 20041/GERM/16).

D. Bautista-Salinas, J. Roca González and O. M. Mozos are with the Technical University of Cartagena, 30202 Cartagena, Spain (corresponding author phone: +34 868071063; fax: +34 968326400; e-mail: oscar.mozos@upct.es).

I. Méndez Mateo is with the University of Murcia, 30100 Espinardo, Murcia, Spain.

very simple to facilitate usage and reduce attention theft (i.e. the shift of the user's attention only to the IT device rather than the "real" environment).The app can be accessed at any time to input the levels too. Happiness and activeness levels are stored with a timestamp using Cloud Firestore (Firebase, Google LLC, CA, USA).

phasic change, the skin conductance response (SCR). We followed the procedure proposed by R. Zangróniz *et al.* [22]. In addition, we extracted features in the time-domain [18], [22] and frequency-domain [19], [22].

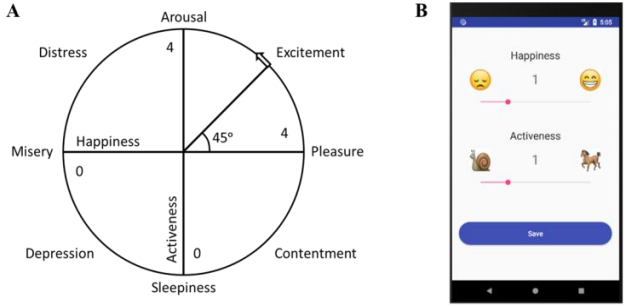

Figure 1. (A) Mental state model [14] ; (B) Mobile application interface.

## A. Physiological Signal Processing and Feature

The different physiological signals are first preprocessed and filtered. Features are extracted using a sliding window approach with window size of 60 seconds. This size has been widely reported as the appropriate value when working with physiological signals [9], [10], [15]. In addition, we set an overlapping of 10% between consecutive windows to reduce the boundary effect that appears when signals are filtered. In Table I we present the 203 features extracted from the following measurements.

A) 3-axis Accelerometer (72 features)

The 3-axis accelerometer signal is sampled at a frequency of 32 Hz. We calculated the Euclidean norm and then we applied a $3^{rd}$ order band-pass Butterworth filter with cut-off frequencies of 0.2 Hz and 10 Hz [16] over the 4 signals. Afterwards, features in the time-domain [16]–[18] and frequency-domain [19] were extracted.

B) Peripheral Skin Temperature (13 features)

The peripheral temperature signal is sampled at a frequency of 4 Hz. We set a threshold of 2 ºC and data were discarded when detecting an increase or decrease greater between consecutive data points. Then, features in the time-domain [18], [20] and frequency-domain [19] features were extracted.

C) HR Variability (27 features)

We cannot reconstruct the IBI signal obtained from the E4 in a reliable way. Therefore, we used the HR file, which is sample at a frequency of 1 Hz. We normalized HR using the first ten minutes from each day as the basal measurement to compare data from different participants. Then, we extract features in the time-domain [16]–[18] and in the frequency-domain [21].

D) EDA (91 features)

The EDA signal is sampled at a frequency of 4 Hz. We applied a $3^{rd}$ order low-pass Butterworth filter with a cut-off frequency of 1.5 Hz [22]. From the EDA signal we extracted its tonic level, the skin conductance level (SCL), and its

TABLE I. FEATURES EXTRACTED FROM THE PHYSIOLOGICAL SIGNALS RECORDED WITH THE EMPATICA E4 DEVICE.

**Accelerometer**

| Time-domain features |
|---|
| Maximum (MAX), 90$^{th}$ percentile (P90), Variance (VAR), Mean Absolute Deviation (MAD) and norm [16]. Amplitude (AMP), Minimum (MIN), Standard Deviation (STD) and Root Mean Square (RMS) [17]. Means of the Absolute Values of the First Differences of the Raw (MAVFD) and Normalize (MAVFDN) Signals. Means of the Absolute Values of the Second Differences of the Raw (MAVSD) and Normalize Signals (MAVSDN) [18]. |
| Frequency-domain features |
| Mean and STD from periodogram P25, P50 and P75 [19]. |

**Temperature**

| Time-domain features |
|---|
| Mean, Slope (SRL) and Intercept (IRL) of a fitted regression line [20]. STD, MAVFD, MAVFDN, MAVSD and MAVSDN [18]. |
| Frequency-domain features |
| Mean and STD from periodogram [19]. P25, P50 and P75. |

**EDA, SCL and SCR**

| Time-domain features |
|---|
| Mean, STD, MAX, MIN, Dynamic Range (DR). First Derivative Mean (FDM) and STD (FDSTD). Second Derivative Mean (SDM) and STD (SDSTD). Arc Length (AL), Integral (I), Normalize Average Power (NAP), Normalized Root Mean Square (NRMS), Area-Perimeter Ratio (APR), Energy-Perimeter Ratio (EPR), Central Moment (CM), Skewness and Kurtosis Statistics [22]. MAVFD, MAVFDN, MAVSD, MAVSDN, SMFD (only for EDA) [18]. |
| Frequency-domain features |
| Mean and STD from periodogram [19]. P25, P50 and P75. Three band powers: [0.1-0.2] Hz, [0.2-0.3] Hz and [0.3-0.4] Hz |

**Heart Rate**

| Time-domain features |
|---|
| MAX, P90, VAR, MAD and norm [16]. Mean, MIN and STD [17]. MAVFD, MAVFDN, MAVSD and MAVSDN [18]. Mean of the Smooth HR (SM) and Mean of the First Differences (MFD) [18]. |
| Frequency-domain features |
| Very Low (aVLF), Low (aLF), High (aHF) and Total Absolute Spectral Power (aTotal). Very Low (pVLF), Low (pLF) and High Frequency Spectral Power in Percentage (pHF). Low (nLF) and High Frequency Spectral Power Normalize to Total Power (nHF). Ratio of Low Frequency to High Frequency (LFHF). Very Low (peakVLF), Low (peakLF) and High Frequency Peak (peakHF) [21]. |

## B. Ground Truth

We use the EMA inputs as ground truth events for training the classifier. Since the number of daily EMA inputs is low (5 times a day) we extend them before and after the moment we receive them. We tried different windows in this work 60, 90 and 120 minutes.

In addition, users usually do not answer the EMA question at the time of activation and sometimes they wait for some period of time before doing it. In case that the gap between EMAs is less than half of the window selected, we will take half of the data for each EMA point. For the first and last EMA of the day, which are often answered within the first and last hour of the day, we will select the amount of data available. Examples of possible scenarios are presented in Fig. 2.

Finally, extended EMA inputs are aligned on time with the physiological signals and the corresponding features, so that we obtained a mood input with its corresponding physiological feature vector of 203 features.

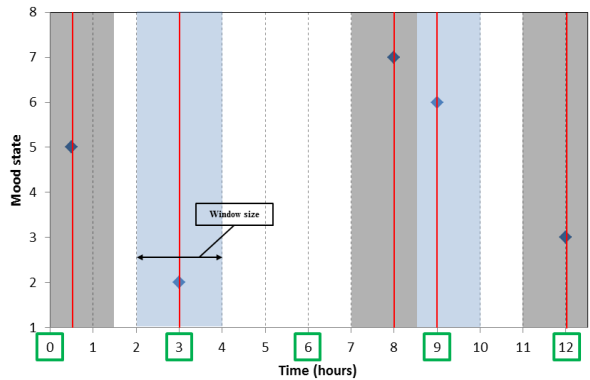

Figure 2. Illustration of the extrapolation of the mood states obtained from the EMA answers received during one day with a 120 minute centered window. Green squares represent the time where the EMAs were programmed. Gray and blue rectangles represent data selected. Red lines represent the time where the EMAs were answered.

### C. Classification

We built a mood classifier using the support vector machine (SVM) library libSVM [23] using a radial basis function (RBF). Parameters C and γ were obtained by grid-search using cross-validation. The feature vectors were scaled in the range [-1, 1].

For classification, we discarded those mood states that represented less than 10% of the total data. The states classified were: pleasure, sleepiness and contentment for participant 1; pleasure, excitement, arousal and contentment for participant 2; excitement, arousal and depression for participant 3; and pleasure, excitement and sleepiness for participant 4. Additionally, we classified the happiness and activeness values from the remaining moods. Two different classification models were used. The first model followed a standard method splitting the data randomly into 75% for training and 25% for testing. In addition, we tried the three EMA windows of 30, 60, and 120 minutes. We repeated the process 5 times and calculated the mean and standard deviation of the accuracy.

To train a second model we use a leave-one(day)-out approach in which we used one complete day for testing and the rest of days to train the classifier. The accuracy was calculated as the mean from all the days classified. The EMA window was the one with better classification results in the first model.

## III. EXPERIMENTS

### A. Data Collection

The study was carried out on participants from the Senior University of Cartagena (n=4). All subjects gave written informed consent in accordance with the Declaration of Helsinki and they were duly informed about the goals of the research. In addition, we carried out personal interviews to collect demographic data and the following questionnaires were used to evaluate their mental health: STAI [24], Hamilton anxiety scale [25], Yesavage geriatric depression scale [26], MEC-35 [27], Ryff scale of psychological well-being [28], global deterioration scale [29], Katz index [30], PANAS [31]. After the interview, the EMA application was installed in their personal smartphone and they received an Empatica E4 device. We collected data from each participant during 15 consecutive days. Throughout this time, participants wore the wristband device during the day hours and removed it when they went to sleep. They also were instructed to answer the daily EMAs sent to their smartphones. At the end of the study, participants came back to the laboratory and repeated the personal interview. They also completed a satisfaction survey and returned the E4 device. According to the questionnaires, participants did not show any symptoms of anxiety, mental disorder or disability. Additionally, the levels for the different questionnaires were maintained stable from the first to the last interview. In Table II, demographical and collected data are summarized.

TABLE II. PARTICIPANTS' DEMOGRAPHICAL AND COLLECTED DATA.

|  | **P1** | **P2** | **P3** | **P4** |
|---|---|---|---|---|
| **Age** | 67 | 55 | 60 | 63 |
| **Gender** | Male | Female | Male | Female |
| **Days** | 9 | 15 | 14 | 14 |
| **Non-valid days** | 0 | 0 | 2 | 1 |
| **Total EMAs (Daily avg.)** | 42 (4.67±0.67) | 57 (3.80±0.75) | 64 (5.33±1.84) | 46 (3.54±0.93) |

### C. Classification Results

We first learned a personalized model of mood for each participant by splitting the corresponding data into 75% for training and 25% for testing and repeating the process 5 times. Fig. 3 shows the classification results for each participant using different EMA windows. In addition, we show the results by using the data from all participants and learning a unique model for all. An ANOVA test was used to test the statistical significance of the results for different EMA windows. Tukey's honest significant difference criterion is used for the multicomparison test.

Figure 4 and 5 shows the classification results when learning a model for the happiness and the activeness. The best accuracies for the 60' window were 90.05%±2.59% (participant 3) for mood classification, 88.93%±0.84% (participant 4) for happiness classification and 87.21%±0.21% (participant 4) for activeness classification.

In a second experiment we used the EMA 60' window (the one with best results) to learn a personalized model using a leave-one(day)-out, in which date from one complete day is used for testing while the data of the remaining days are used for training. The resulting mood accuracy rates were below 40% (Fig. 6). This is a low accuracy in comparison with the previous experiment, however, results make sense since we do not have similar examples in the training data and the testing day is completely new.

To check the correctness of our results using leave-one(day)-out, we compare with the study by R. Likamwa *et al.* [13], which is the closest to ours. For this comparison we learn a model for the happiness which is the dimension used in [13]. Our learned model presents a similar performance to [13] as shown in Fig. 7. Moreover, the study in [13] uses periods of 10 and 20 days. We use 15 days and as shown in the plot, our accuracies lay mostly in between. Unfortunately, the work in [13] does not present results of mood prediction as we do in this paper.

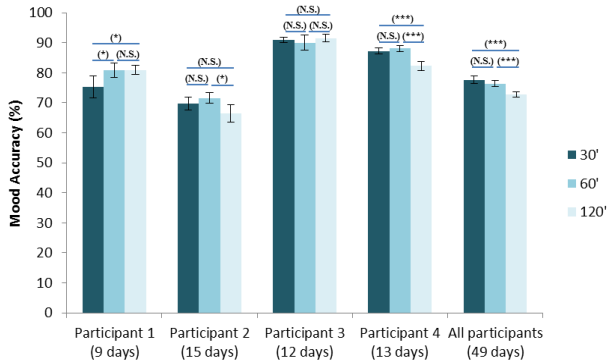

Figure 3. Bar plot showing the system's mood accuracy for the 30 m, 60 m and 120 m windows. Error bars represent the standard deviation. (*): Significant, P<0.05. (***): Significant, P<0.001. N.S.: Not significant.

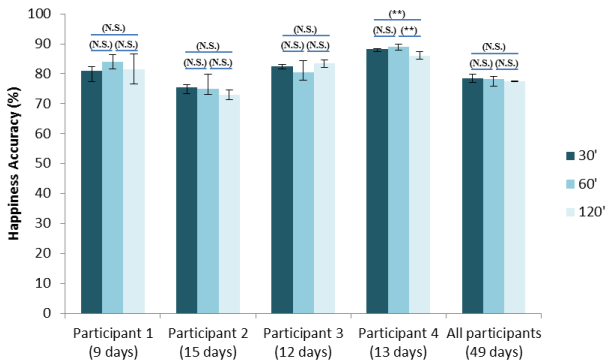

Figure 4. Bar plot showing the system's happiness accuracy for the 30 m, 60 m and 120 m windows. Error bars represent the standard deviation. (**): Significant, P<0.01. N.S.: Not significant.

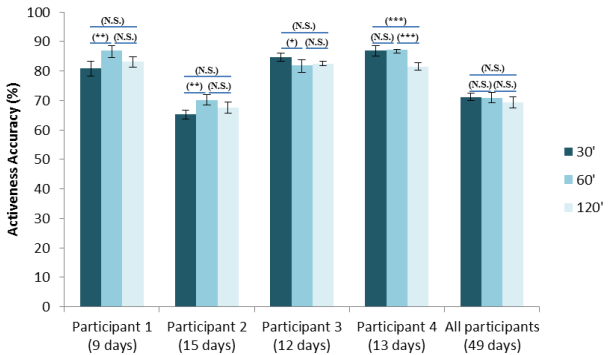

Figure 5. Bar plot showing the system's activeness accuracy for the 30 m, 60 m and 120 m windows. Error bars represent the standard deviation. (*): Significant, P<0.05. (**): Significant, P<0.01. (***): Significant, P<0.001. N.S.: Not significant.

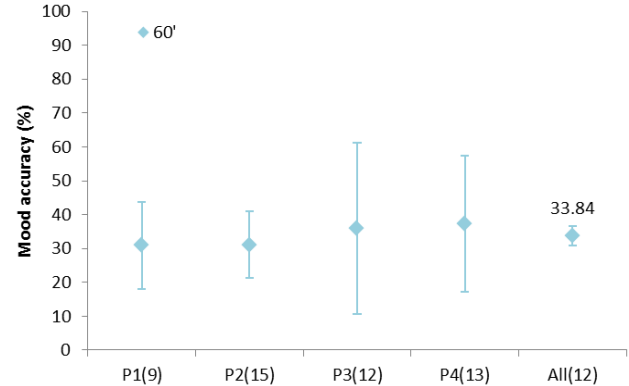

Figure 6. Plot showing the system's mood accuracy for the 60' window using the classification method leave-one(day)-out. Error bars represent the standard deviation.

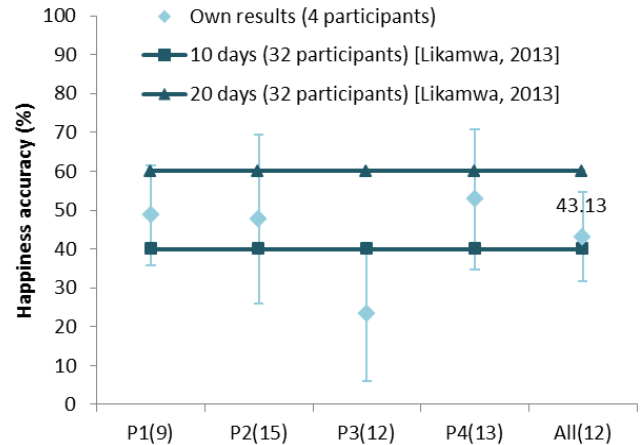

Figure 7. Plot showing the system's happiness accuracy for the 60' window using the classification method leave-one(day)-out. Error bars represent the standard deviation. Square and Triangle dots represent the mean happiness accuracy for datasets of 10 and 20 days respectively reported by R. Likamwa *et al.* [13].

## IV. DISCUSSION AND CONCLUSIONS

We found that the EMA 60' window works significantly better for the mood recognition system proposed. Moreover, the accuracy rates for mood and happiness and activeness levels are similar or higher than those reported in the literature [13], [32], [33]. Although we cannot assure that there are not mood states changes in some of the windows selected, in comparison with other works that take a small window of ±10 m [20], our approach clearly takes advantage of the data collected in a better way.

The accuracy rates for the method leave-one(day)-out are in accordance with other studies using the same classification method and number of days [13]. However, it has been reported that between 40 and 60 days are needed to achieve accuracy rates above 80% in happiness level prediction [13]. Another study suggests that at least 20-25 days might be necessary in order to build a person-specific mood classifier [34]. Therefore, we plan to extend the number of days in our future experiments. Right now, we only use the physiological data for predicting the mood, but other sources of information can be added like the patterns of the use of the mobile phone, or the activity agenda of participants (when available). A higher number of participants is also needed to test generalization and transferability of the mood models. Moreover, deep learning could be used as a classifier in future experiments when bigger amount of data is available.

Acquiring ground truth for supervised learning is an open problem. We decided for a very simple app to ask mood states to participants 5 times a day, although participants could voluntarily increase the number of inputs. This number was suggested by psychologists. The last row in Table II shows that participants averaged 4-5 times a day which means this is a viable number.

The experimental data in this work was collected during daily life activities of people with no restrictions on their behavior. The participants did no find intrusion in their lives, which shows that our technology is viable for daily life use.

ACKNOWLEDGMENTS

We would like to thank all the participants for taking part in the study.